\documentclass[11pt]{article}

\usepackage[final]{acl}

\usepackage{times}
\usepackage{latexsym}
\usepackage[T1]{fontenc}
\usepackage[utf8]{inputenc}
\usepackage{inconsolata}
\usepackage{microtype}
\usepackage{graphicx}
\usepackage{booktabs}
\usepackage{colortbl}
\usepackage{amsmath}
\usepackage{amssymb}
\usepackage{multirow}
\usepackage{algorithm}
\usepackage{algpseudocode}
\usepackage{tikz}
\usetikzlibrary{arrows.meta,positioning,calc}
\usepackage{placeins}
\usepackage{dblfloatfix}
\usepackage{balance}
\graphicspath{{figures/}}

\newcommand{\method}{\textsc{TRAC}}

\newcommand{\llada}{\textsc{LLaDA}}

\title{TRAC: Efficient Diffusion Language Models for Machine Translation\\ via Length-Aware Decoding}

\author{
  Tianyuan Li\textsuperscript{1}, Xihan Yang\textsuperscript{1}, Chenglong Wang\textsuperscript{1} \\
  \textbf{Yuzhang Wu\textsuperscript{1}, Jingbo Zhu\textsuperscript{1,2}, Tong Xiao\textsuperscript{1,2}\thanks{Corresponding author.}} \\
  \textsuperscript{1}School of Computer Science and Engineering, Northeastern University, Shenyang, China \\
  \textsuperscript{2}NiuTrans Research, Shenyang, China \\
    \texttt{lity3@mails.neu.edu.cn} \\
  \texttt{\{zhujingbo,xiaotong\}@mail.neu.edu.cn}
}
\begin{document}

\maketitle

\begin{abstract}
Machine translation with large language models requires both high translation quality and low generation latency. Accelerating translation while preserving quality is therefore essential for practical deployment. Diffusion large language models (dLLMs) can generate tokens at arbitrary target positions through iterative denoising, offering a new paradigm for parallel translation beyond sequential autoregressive decoding. However, existing methods still incur redundant computation: fixed target generation lengths waste computation on unused positions, while confidence-only commitment delays medium-confidence tokens whose predictions have largely stabilized. To address these issues, we propose TRAC, a training-free adaptive decoding method for diffusion machine translation. TRAC adapts the target generation length using source length and translation direction, with a conservative fallback to the maximum length for empty or unusually short outputs. It preserves high-confidence commitment while identifying temporally stable medium-confidence tokens for early commitment using cross-step top-1 persistence and approximate Jensen–Shannon divergence over Top-$K$ distributions. On bidirectional Chinese–English translation across FLORES-200, WMT23, and NTREX-128, TRAC reduces model forward calls by 24.9\%–36.8\% and achieves 1.91–2.19× sentence-level generation speedups over Fast-dLLM, with only a 0.02-point decrease in macro-average COMET. These results demonstrate that adaptive generation lengths and stability-aware token commitment improve the efficiency of diffusion machine translation.
\end{abstract}

\section{Introduction}
\label{sec:introduction}

Machine translation with large language models (LLMs) requires both high translation quality and low sentence-level generation latency. Most LLMs still use autoregressive (AR) decoding, generating translations one token at a time. This sequential dependency limits parallel generation and creates a major bottleneck in translation inference.

Diffusion language models (dLLMs) provide a promising alternative by generating multiple target positions in parallel through iterative denoising~\citep{austin2021d3pm,sahoo2024mdlm,nie2025llada}. 
In masked dLLMs such as \llada{}, decoding first allocates a fixed number of target token positions, which we refer to as the \emph{target canvas}, and then performs a fixed number of decoding steps determined by the canvas size.
However, parallel prediction does not automatically yield faster translation: an oversized canvas introduces unused positions, while stable but uncommitted tokens continue to participate in subsequent iterations, incurring additional computation. To identify the sources of this redundancy, we systematically analyze dLLM decoding behavior and prediction trajectories in bidirectional Chinese--English translation, yielding two key observations.
\begin{figure*}[t]
    \centering
    \includegraphics[width=\textwidth]{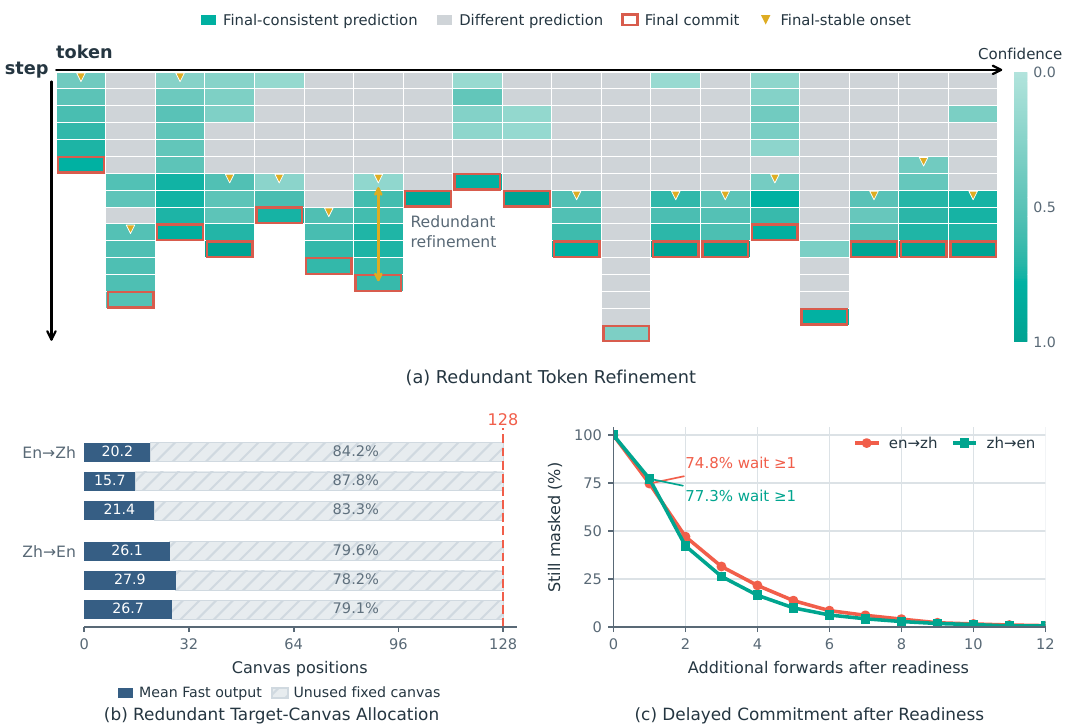}
    \caption{Spatial and temporal redundancy in diffusion machine translation. (a) A stepwise Fast-dLLM decoding trajectory: each row represents one model forward pass, which predicts all target positions simultaneously. Color intensity encodes prediction confidence. (b) The fixed 128-token canvas substantially exceeds the effective output length across translation settings. (c) A trajectory audit of 400 FLORES samples reveals many positions that satisfy temporal convergence conditions but remain masked and participate in subsequent forward passes.}
    \label{fig:motivation}
\end{figure*}
First, a fixed target canvas incurs substantial unnecessary computation. With a maximum generation length of $128$, only $12.2\%$--$21.8\%$ of the canvas is used on average across our evaluation settings. Second, commitment based on single-step confidence leads to repeated iterations. Many medium-confidence tokens remain uncommitted even after their predictions have become stable across consecutive denoising steps, causing them to repeatedly participate in later forward passes.

These observations reveal two mismatches underlying redundant dLLM translation computation: at the sentence level, the fixed canvas does not match the actual translation length; at the token level, commitment timing does not match temporal stability. 

To address these issues, we propose \textbf{TRAC}, a training-free adaptive decoding framework for diffusion machine translation. TRAC allocates the canvas using the source--target length relationship associated with source length and translation direction, with a conservative fallback for insufficient space. It further builds a dual-channel token commitment policy on the prefix KV cache, combining the original high-confidence commitment path with a stability-aware path that selectively commits medium-confidence tokens exhibiting cross-step temporal stability.
Figure~\ref{fig:motivation} summarizes the central idea: TRAC reduces redundant computation through canvas allocation before decoding and token commitment during iterative denoising.

Experiments on bidirectional Chinese–English translation across FLORES-200, WMT23, and NTREX-128 show that TRAC reduces model forward calls by $24.9\%$--$36.8\%$ over Fast-dLLM and achieves $1.91$--$2.19\times$ sentence-level generation speedups, with only a $0.02$-point decrease in macro-average COMET. These results show that translation-aware canvas allocation and token commitment informed by temporal stability improves the efficiency of diffusion machine translation.

\section{Related Work}

\subsection{Diffusion Large Language Models}

Diffusion large language models (dLLMs) replace autoregressive generation with iterative denoising, enabling bidirectional modeling and parallel updates across multiple positions~\citep{austin2021d3pm,sahoo2024mdlm}. Recent masked dLLMs, including LLaDA and Dream, have scaled to billions of parameters and achieved competitive performance across a range of language understanding and generation tasks~\citep{nie2025llada,ye2025dream}. Standard LLaDA uses semi-autoregressive decoding: the target sequence is partitioned into blocks generated sequentially from left to right, while each block undergoes multiple denoising steps that predict all masked positions in parallel and progressively commit tokens according to prediction confidence. Although this strategy enables parallel generation within each block, its fixed, predefined number of denoising steps per block does not readily adapt to differences in translation length across inputs or in convergence speed across tokens, resulting in unnecessary computation.

\subsection{Efficient Decoding for Diffusion Language Models}

Inference efficiency remains a major bottleneck for dLLMs because high-quality generation typically requires multiple Transformer forward passes. Existing methods primarily reduce this overhead through cache reuse, attention sparsification, and more aggressive parallel decoding~\citep{long2026focus}. Fast-dLLM introduces both an approximate KV cache for dLLMs and confidence-based parallel decoding, allowing multiple high-confidence positions to be committed in the same step~\citep{wu2025fastdllm}. We adopt Fast-dLLM's prefix KV cache as our inference foundation and focus on further optimizing the decoding schedule.

A growing body of work uses decoding dynamics to adaptively determine when to commit tokens. CadLLM adjusts the decoding budget based on confidence changes during denoising~\citep{shen2026cadllm}; Ada-DLM uses cross-step confidence trajectories to identify stable positions that static thresholds fail to capture promptly~\citep{li2026adadlm}; and CreditDecoding accumulates evidence from previous predictions and recalibrates the current logits, allowing low-confidence tokens with persistent support to meet the commitment criterion earlier~\citep{wang2026credit}. LocalLeap, Prophet, and LESS also reduce redundant refinement through local certainty propagation, early commitment, and joint cross-step stability, respectively~\citep{kong2025localleap,li2025prophet,mohamed2026less}. Recent work has also explored reinforcement-learning-based post-training for dLLMs, improving reasoning capability while learning more efficient inference behaviors, such as adaptive sampling or reduced denoising budgets~\citep{wang2025mro,zhao2025diffpo}.

In contrast to these approaches, TRAC does not modify the current logits or replace the original confidence-based decoding mechanism with a stability rule. We retain Fast-dLLM's direct high-confidence commitment path and use cross-step stability only for medium-confidence positions that have not reached the high threshold. This recovers additional opportunities for parallel commitment without increasing decision overhead for high-confidence tokens.

\begin{figure*}[t]
    \centering
    \includegraphics[width=\linewidth,keepaspectratio]{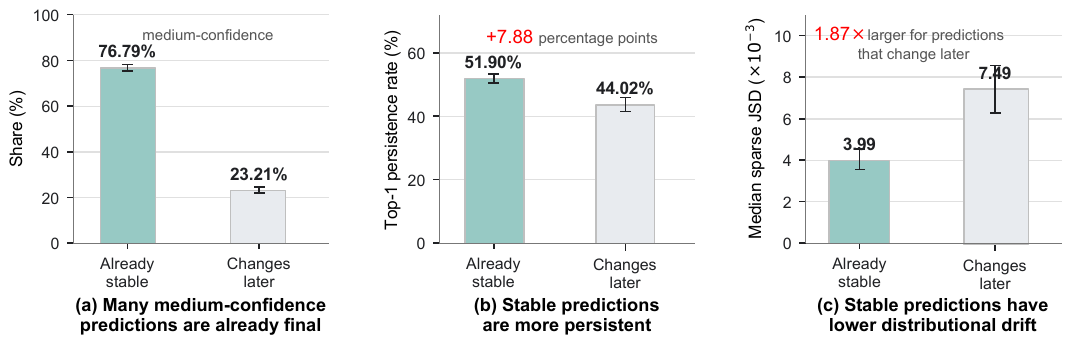}
    \caption{
Temporal analysis of medium-confidence predictions under Fast-dLLM.
(a) Most observations are already stable.
(b) Already-stable observations exhibit greater top-1 persistence.
(c) They also exhibit lower cross-step sparse JSD.
Labels are assigned retrospectively using the complete baseline trajectory and are used only for analysis. Error bars denote 95\% confidence intervals from 2,000 sentence-level bootstrap resamples stratified by translation direction.
}
    \label{fig:temporal-analysis}
\end{figure*}

\subsection{Parallel and Non-Autoregressive Machine Translation}


Machine translation research includes efficient systems~\citep{wang2021niutrans}, reinforcement-learning training~\citep{wang2024esrl}, structure-aware document translation~\citep{zhu2026latextrans}, and parallel generation. Non-autoregressive machine translation (NAT) removes the sequential dependencies of autoregressive decoding by predicting target tokens in parallel, but faces challenges including target multimodality, length prediction, and quality degradation~\citep{gu2018nat,ghazvininejad2019mask,schmidt2022clarity}. Diff-DAT combines a diffusion process with a Directed Acyclic Transformer, offering an adjustable quality--latency trade-off while retaining parallel generation~\citep{nguyentri2025diffdat}.

Unlike autoregressive models, which can terminate naturally by generating EOS, masked diffusion translation typically requires the target space to be specified before denoising begins. Target-length or canvas selection therefore introduces an additional inference decision. Entropy-Valley compares the prediction entropy of multiple fully masked candidate canvases to select a target length better suited to denoising for each input. It also shows that a diffusion-friendly canvas need not exactly match the reference translation length~\citep{zhan2026entropyvalley}.

TRAC also addresses target canvas allocation, but aims to reduce inference computation rather than improve length prediction accuracy. Without enumerating candidate canvases or introducing additional model forward passes, we directly allocate block-aligned decoding space using direction-specific relationships between source and target lengths in machine translation. We view this space as a diffusion workspace that includes the slack needed for refinement, rather than as a direct prediction of the reference translation length.

\section{Motivating Analysis}

Before introducing our method, we analyze Fast-dLLM's decoding behavior in machine translation. We focus on two factors that directly affect inference cost: the target space allocated before denoising and the repeated refinement that tokens undergo before final commitment. Figures~\ref{fig:motivation} and~\ref{fig:temporal-analysis} present representative evidence for these two factors, respectively.

\subsection{Target-Space Redundancy}

\label{sec:target-space-redundancy}

Masked diffusion decoding requires the target canvas to be specified before denoising begins, and standard settings typically allocate the same maximum space to every input. With a fixed canvas of $C=128$, Fast-dLLM achieves average canvas utilization of only $14.91\%$ for En$\rightarrow$Zh and $21.02\%$ for Zh$\rightarrow$En, indicating that many target positions do not appear in the final translation.

However, the required target space is not entirely unpredictable. On the development set, the correlation between source and target token lengths is approximately $0.84$ in both directions, while En$\rightarrow$Zh and Zh$\rightarrow$En exhibit different length ratios. These observations suggest that a fixed maximum canvas is overly conservative for machine translation, and that source sentence length and translation direction provide an inexpensive prior for allocating target space to each instance.

\subsection{Temporal Refinement Redundancy}

\label{sec:temporal-redundancy}

Fast-dLLM primarily determines whether to commit a token based on its current confidence, so positions below the high-confidence threshold continue to undergo refinement in subsequent steps. We record decoding trajectories from unmodified Fast-dLLM and focus on medium-confidence observations satisfying $0.5 \leq c_i^{(t)} < 0.8$.

We analyze the complete baseline trajectories retrospectively. We label an observation as \emph{already stable} if its current top-1 prediction remains unchanged throughout the rest of the Fast-dLLM trajectory and is ultimately committed at that position; otherwise, we label it as \emph{changes later}. We find that $76.79\%$ of medium-confidence observations already have a top-1 prediction that remains consistent with the token ultimately committed at that position. In other words, the top-1 predictions of many positions that have not yet passed the high-confidence gate have already stopped changing.

We further examine whether cross-step history can identify these latent stable states. As shown in Figure~\ref{fig:temporal-analysis}, $51.90\%$ of \emph{already stable} observations retain the same top-1 prediction across adjacent steps, compared with $44.02\%$ of \emph{changes later} observations. This suggests that persistence provides some, albeit limited, discriminative information. Among positions with an unchanged top-1 prediction across adjacent steps, the median sparse JSD is $0.00399$ for \emph{already stable} observations and $0.00749$ for \emph{changes later} observations; the latter is approximately $1.87\times$ the former.

These results show that instantaneous confidence and temporal convergence are not perfectly synchronized. Rather than lowering the confidence threshold uniformly, a more natural approach is to preserve Fast-dLLM's existing high-confidence fast path and use cross-step stability as additional evidence only for medium-confidence positions.
\section{Method}
\label{sec:method}

\subsection{Overview}
We propose TRAC, a training-free adaptive decoding framework for diffusion machine translation that addresses the target-space and iterative-refinement redundancies identified above. Existing methods use a fixed target canvas and rely primarily on single-step confidence to decide which tokens to commit. TRAC instead organizes diffusion translation decoding around two complementary adaptive decisions: target-space allocation before denoising and early token commitment during denoising.

\begin{figure*}[t]
    \centering
    \includegraphics[width=\textwidth]{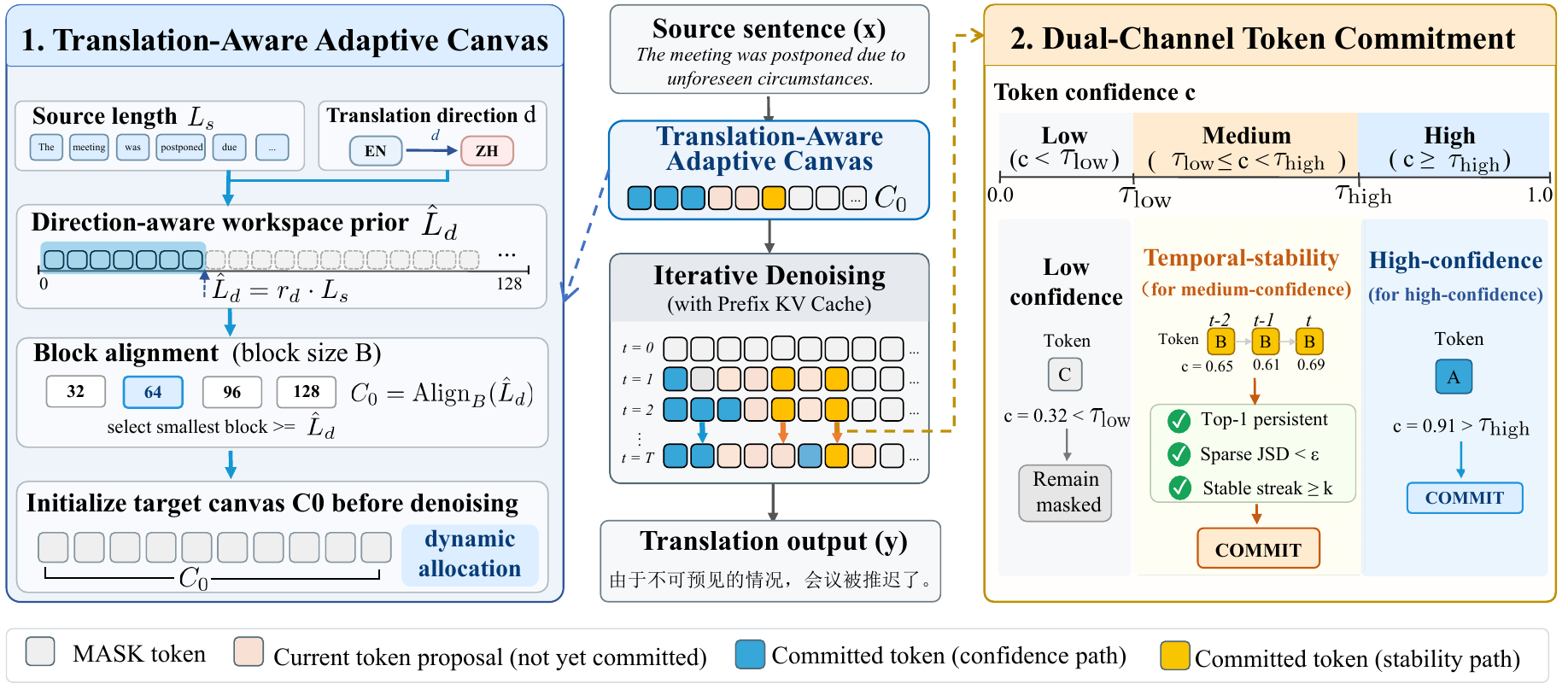}
    \caption{Overview of TRAC. Given a source sentence, TRAC first allocates a translation-aware adaptive canvas according to source length and translation direction, then performs iterative denoising with dual-channel token commitment to commit both high-confidence and temporally stable tokens.}
    \label{fig:method-overview}
\end{figure*}

TRAC comprises two core modules. Translation-Aware Adaptive Canvas exploits the relationship between source and target lengths in machine translation to dynamically allocate target generation space according to source length and translation direction, reducing unnecessary computation from a large fixed canvas. Dual-Channel Token Commitment preserves Fast-dLLM's original high-confidence path and adds a second commitment channel based on temporal stability for medium-confidence positions, allowing positions approaching convergence to exit subsequent denoising earlier. The method requires only lightweight changes to the decoding schedule and is fully compatible with the prefix KV cache. As shown in Figure~\ref{fig:method-overview}, the two modules govern space allocation before denoising and token commitment at each step, respectively.

\subsection{Translation-Aware Target Canvas Allocation}
\label{sec:adaptive-canvas}

Motivated by the target-space redundancy observed in Section~\ref{sec:target-space-redundancy}, we adaptively allocate a target canvas for each input according to source length and translation direction, providing an appropriate diffusion decoding workspace for the instance.

Given a source sentence $x$, let $L_s$ denote its token count under the LLaDA tokenizer and $d$ the translation direction. We first estimate the base target-space requirement:
\begin{equation}
\hat{L}(x,d) = r_d L_s,
\label{eq:length-prior}
\end{equation}
where $r_d$ is a fixed direction-specific ratio.

The target canvas reserves refinement workspace for iterative denoising. We therefore align the allocation with Fast-dLLM's block structure instead of directly using $\hat{L}(x,d)$.

We first constrain the base length estimate to the allowed canvas range:
\begin{equation}
\bar{L}(x,d)=\operatorname{clip}\!\left(
\operatorname{round}(\hat{L}(x,d)), C_{\min}, C_{\max}
\right),
\label{eq:clip-canvas}
\end{equation}
and then round it up to a multiple of Fast-dLLM's block length $B$:
\begin{equation}
C_0(x,d)=\min\!\left\{
C_{\max},
B\left\lceil\frac{\bar{L}(x,d)}{B}\right\rceil
\right\}.
\label{eq:block-aligned-canvas}
\end{equation}
Here, $C_{\min}$ and $C_{\max}$ control the minimum and maximum allowed target space, respectively, and $B$ is the block length. Rounding up reduces the cost of a fixed maximum canvas for shorter inputs while retaining some additional space, rather than equating the diffusion refinement workspace exactly with the predicted translation length.

\paragraph{Conservative Fallback.}
The direction-specific length prior cannot cover every instance, and the dynamic canvas may still under-allocate space for a small number of examples. We therefore introduce a conservative fallback mechanism.

The model first decodes with the dynamic canvas $C_0(x,d)$. Let $y^{(1)}$ denote the first translation and $L_{\mathrm{out}}$ its effective output length. If the translation is empty or satisfies
\begin{equation}
L_{\mathrm{out}} < \alpha\hat{L}(x,d),
\label{eq:fallback-condition}
\end{equation}
the fallback is triggered, and decoding restarts from the initial state with the full canvas $C_{\max}$. Here, $\alpha$ is the short-output detection threshold. We retain the second decoding result and include the latency and NFE of both decoding runs in the reported experimental cost.

This fallback decision relies only on the source-length prior and the model's own output. It does not use reference translations, an additional length-prediction model, or an external quality evaluator.

\subsection{Dual-Channel Convergence-Aware Commitment}
\label{sec:dual-channel}

Motivated by the observation in Section~\ref{sec:temporal-redundancy}, we retain the original high-confidence commitment rule and introduce temporal stability only for medium-confidence positions, yielding two parallel token acceptance channels.

For a position $i\in\mathcal{M}_t$ that remains masked at denoising step $t$, the model produces a predictive distribution $p_i^{(t)}$. We define its top-1 proposal and corresponding confidence as
\begin{equation}
\hat{y}_i^{(t)}=\arg\max_{v\in\mathcal{V}}p_i^{(t)}(v),
\qquad
c_i^{(t)}=\max_{v\in\mathcal{V}}p_i^{(t)}(v).
\label{eq:proposal-confidence}
\end{equation}
Given $\tau_{\mathrm{low}}<\tau_{\mathrm{high}}$, the set of medium-confidence positions is defined as
\begin{equation}
\mathcal{M}_t^{\mathrm{mid}}=
\left\{i\in\mathcal{M}_t:
\tau_{\mathrm{low}}\leq c_i^{(t)}<\tau_{\mathrm{high}}
\right\}.
\label{eq:medium-set}
\end{equation}
Positions at or above $\tau_{\mathrm{high}}$ enter the original high-confidence path; positions in $\mathcal{M}_t^{\mathrm{mid}}$ undergo temporal stability checks; and positions below $\tau_{\mathrm{low}}$ remain masked and participate in subsequent refinement.

\paragraph{High-Confidence Fast Path.}
The set of high-confidence positions is
\begin{equation}
\mathcal{A}_t^{\mathrm{high}}
=\left\{i\in\mathcal{M}_t:
c_i^{(t)}\geq\tau_{\mathrm{high}}\right\}.
\label{eq:high-channel}
\end{equation}
We commit these positions directly using Fast-dLLM's original rule.

\paragraph{Medium-Confidence Stability Path.}
For positions in $\mathcal{M}_t^{\mathrm{mid}}$, we combine confidence thresholds with predictive stability. We first require top-1 persistence across two consecutive steps:
\[
\hat{y}_i^{(t)}
=
\hat{y}_i^{(t-1)}.
\]
We then compare the predictive distributions across these steps using Jensen--Shannon divergence. For two probability distributions $P$ and $Q$, their divergence is defined as~\citep{lin1991jsd}
\begin{equation}
\begin{aligned}
M = & \frac{P+Q}{2},\\
D_{\mathrm{JS}}(P,Q)
=& \frac{1}{2}D_{\mathrm{KL}}(P\|M) + \\
&\frac{1}{2}D_{\mathrm{KL}}(Q\|M).
\end{aligned}
\label{eq:jsd}
\end{equation}

We consider position $i$ stable at the current step if it remains in the medium-confidence interval across consecutive steps, retains the same top-1 prediction, and satisfies
\[
D_{\mathrm{JS}}
\left(
\tilde{p}_i^{(t-1)},
\tilde{p}_i^{(t)}
\right)
<
\epsilon
\]
using the sparse distributions defined below.

To avoid premature commitment due to a single instance of chance agreement, we further require stability to persist for multiple consecutive steps. Let $r_i^{(t)}$ denote the number of consecutive steps for which position $i$ satisfies the stability conditions above, with a newly observed medium-confidence position assigned a streak length of 1.
We update it as follows:
\begin{equation}
r_i^{(t)}=
\begin{cases}
0, & i\notin\mathcal{M}_t^{\mathrm{mid}},\\
r_i^{(t-1)}+1, & \text{if $i$ is stable at step $t$},\\
1, & \text{otherwise}.
\end{cases}
\label{eq:stable-streak}
\end{equation}

Once the number of consecutive stable steps reaches the threshold $k$, the position enters the stability-based commitment channel:
\begin{equation}
\mathcal{A}_t^{\mathrm{stable}}=
\left\{i\in\mathcal{M}_t^{\mathrm{mid}}:r_i^{(t)}\geq k\right\}.
\label{eq:stable-channel}
\end{equation}

The final commitment set at the current step is the union of the high-confidence and stability-based channels:
\begin{equation}
\mathcal{A}_t=
\mathcal{A}_t^{\mathrm{high}}\cup\mathcal{A}_t^{\mathrm{stable}}.
\label{eq:dual-channel}
\end{equation}

\paragraph{Efficient Sparse Stability Estimation.}

Storing full-vocabulary distributions across consecutive denoising steps would incur substantial additional memory costs. For medium-confidence positions, we therefore use the union of the top-$K$ candidates from the two steps as a shared support set. At each step, probability mass outside that step's top-$K$ is aggregated into an \texttt{OTHER} category, and JSD is computed on the resulting low-dimensional representations.

Specifically, let $S_i^{(t-1)}$ and $S_i^{(t)}$ denote the top-$K$ candidate sets at position $i$ at steps $t-1$ and $t$, respectively. We form
\begin{equation}
U_i^{(t)}
=
S_i^{(t-1)}
\cup
S_i^{(t)},
\end{equation}
and align the predictions from both steps over this set. For a candidate outside a given step's top-$K$, we do not retain its probability individually; instead, the total probability mass of all unretained candidates is aggregated into \texttt{OTHER}. Thus, each step's sparse distribution contains at most $2K+1$ categories.

Using the resulting sparse distributions $\tilde p_i^{(t-1)}$ and $\tilde p_i^{(t)}$, we compute the sparse approximate Jensen--Shannon divergence
\[
D_{\mathrm{JS}}
\left(
\tilde p_i^{(t-1)},
\tilde p_i^{(t)}
\right)
\]
to estimate the distributional change between consecutive steps.

To further reduce overhead, JSD is computed only for positions that remain in the medium-confidence interval over two consecutive steps and retain the same top-1 prediction. For a current block of length $B$, the additional state retained across steps requires only $O(BK)$ space, whereas storing the full-vocabulary distribution history would require $O(B|\mathcal V|)$. All stability information comes from logits already produced by the current forward pass, requiring no additional Transformer forward passes.

If neither channel yields a position to commit, we follow Fast-dLLM's fallback rule and commit the highest-confidence remaining masked position to ensure that decoding continues to make progress.

\paragraph{Overall Decoding Procedure.}
Algorithm~\ref{alg:trac-decoding} summarizes the complete TRAC decoding procedure. TRAC first allocates a block-aligned target canvas from the source length and translation direction, and then performs iterative denoising using the two commitment channels defined above. If the initial canvas produces an empty or unusually short output, decoding is restarted with the maximum canvas as a conservative fallback.

\begin{algorithm}[t]
\algrenewcommand{\algorithmicindent}{1em}
\caption{Overall decoding procedure of TRAC.}
\label{alg:trac-decoding}
\begin{algorithmic}[1]
\Require source $x$, direction $d$
\State $L_s \gets \operatorname{TokenLength}(x)$
\State $\hat{L} \gets r_d L_s$
\State $C \gets \operatorname{BlockAlign}
    (\operatorname{clip}(\operatorname{round}(\hat{L}),
    C_{\min},C_{\max}), B)$
\State $y \gets \operatorname{Decode}(x,C)$
\If{$y$ is empty \textbf{or} $|y| < \alpha\hat{L}$}
    \State $y \gets \operatorname{Decode}(x,C_{\max})$
\EndIf
\State \Return $y$

\Function{Decode}{$x,C$}
    \State Initialize $C$ masked target positions
    \For{each target block}
        \State Initialize temporal states
        \While{masked positions remain}
            \State Run one Fast-dLLM forward pass with prefix KV cache
            \State $\mathcal{A}^{\mathrm{high}}
            \gets \{i:c_i^{(t)}\geq\tau_{\mathrm{high}}\}$
            \State Identify $\mathcal{A}^{\mathrm{stable}}$ from medium-confidence
            positions
            \Statex \hspace{\algorithmicindent}
            using top-1 persistence, sparse JSD, and stability streak
            \State $\mathcal{A}
            \gets \mathcal{A}^{\mathrm{high}}
            \cup \mathcal{A}^{\mathrm{stable}}$
            \If{$\mathcal{A}=\varnothing$}
                \State $\mathcal{A}\gets
                \{\arg\max_i c_i^{(t)}\}$
            \EndIf
            \State Commit $\hat{y}_i^{(t)}$ for all $i\in\mathcal{A}$
        \EndWhile
    \EndFor
    \State \Return decoded output
\EndFunction
\end{algorithmic}
\end{algorithm}

\section{Experiments}

\definecolor{tracbg}{RGB}{244,248,252}
\definecolor{tracspeedup}{RGB}{34,96,145}
\definecolor{tracgroupsep}{RGB}{210,216,222}

\begin{table*}[!t]
\centering
\small
\setlength{\tabcolsep}{2.8pt}
\renewcommand{\arraystretch}{1.08}
\newcommand{\traclatency}[2]{%
  \begin{tabular}[c]{@{}c@{}}#1\\[-0.35ex]
  {\scriptsize(\textcolor{tracspeedup}{\bfseries #2$\times$})}\end{tabular}}
\begin{tabular}{llccccccc}
\toprule
Model / Decoder
& Metric
& \multicolumn{2}{c}{FLORES}
& \multicolumn{2}{c}{WMT23}
& \multicolumn{2}{c}{NTREX}
& \begin{tabular}[c]{@{}c@{}}Macro\\Avg.\end{tabular} \\
\cmidrule(lr){3-4}\cmidrule(lr){5-6}\cmidrule(lr){7-8}
& & En$\rightarrow$Zh & Zh$\rightarrow$En
& En$\rightarrow$Zh & Zh$\rightarrow$En
& En$\rightarrow$Zh & Zh$\rightarrow$En & \\
\midrule
Qwen2.5-7B
& COMET $\uparrow$ & 87.62 & 86.80 & 85.00 & 79.74 & 84.93 & 84.19 & 84.71 \\
& Latency (s) $\downarrow$ & 0.646 & 0.706 & \textbf{0.517} & 0.805
& \textbf{0.671} & 0.716 & 0.677 \\
\addlinespace[2pt]
Llama-3-8B
& COMET $\uparrow$ & 82.08 & 85.57 & 79.74 & 78.62 & 78.98 & 82.70 & 81.28 \\
& Latency (s) $\downarrow$ & 0.892 & 0.830 & 0.681 & 0.896 & 0.914 & 0.835 & 0.841 \\
\arrayrulecolor{tracgroupsep}
\cmidrule(lr){1-9}
\arrayrulecolor{black}
Vanilla \llada{}
& COMET $\uparrow$ & 86.54 & 84.87 & 84.34 & 77.95 & 83.01 & 83.05 & 83.29 \\
& Latency (s) $\downarrow$ & 9.235 & 8.905 & 8.712 & 8.867 & 9.244 & 9.124 & 9.014 \\
\addlinespace[2pt]
Fast-dLLM
& COMET $\uparrow$ & 85.97 & 84.69 & 83.40 & 77.72 & 82.12 & 82.54 & 82.74 \\
& Latency (s) $\downarrow$ & 1.232 & 0.973 & 1.062 & 1.215 & 1.321 & 1.069 & 1.145 \\
\addlinespace[2pt]
\rowcolor{tracbg}
\method{} \textbf{(Ours)}
& COMET $\uparrow$ & 85.21 & 84.98 & 83.42 & 78.57 & 81.96 & 82.19 & 82.72 \\
\rowcolor{tracbg}
& \begin{tabular}[c]{@{}l@{}}Latency (s) $\downarrow$\\[-0.3ex]{\scriptsize vs. Fast-dLLM}\end{tabular}
& \traclatency{\textbf{0.616}}{2.00}
& \traclatency{\textbf{0.444}}{2.19}
& \traclatency{0.536}{1.98}
& \traclatency{\textbf{0.637}}{1.91}
& \traclatency{0.673}{1.96}
& \traclatency{\textbf{0.535}}{2.00}
& \traclatency{\textbf{0.573}}{2.00} \\
\bottomrule
\end{tabular}
\caption{Main results across six dataset--direction settings. Each cell reports COMET (top) and sentence-level latency in seconds (bottom). For \method{}, parentheses indicate speedup over Fast-dLLM.}
\label{tab:main-results}
\end{table*}

\subsection{Experimental Setup}
\paragraph{Datasets.}

We evaluate English-to-Chinese (En$\rightarrow$Zh) and Chinese-to-English (Zh$\rightarrow$En) translation on FLORES-200~\citep{nllb2022}, WMT23~\citep{kocmi2023wmt}, and NTREX-128~\citep{federmann2022ntrex}. We combine the FLORES dev and devtest sets; for NTREX-128, we construct the reverse Zh$\rightarrow$En setting by swapping the source and target texts. Newstest2020 serves as the development set, and all decoding parameters are selected during development and fixed for final testing. Canvas allocation and fallback detection do not use reference translations at test time.

\paragraph{Models and Baselines.}

All diffusion-model experiments use LLaDA-8B-Instruct~\citep{nie2025llada} and compare Vanilla LLaDA, Fast-dLLM, and TRAC. Vanilla LLaDA uses a fixed 128-token target canvas. Fast-dLLM uses a prefix KV cache and confidence-based parallel decoding and serves as our primary acceleration baseline~\citep{wu2025fastdllm}. We select $\tau=0.8$ on the development set and fix it for all final tests. TRAC is built on the same Fast-dLLM inference framework and does not modify model parameters or the caching mechanism. We also compare two similarly sized autoregressive baselines, Qwen2.5-7B-Instruct~\citep{qwen25} and Llama-3-8B-Instruct~\citep{llama3}, both using greedy decoding and a standard KV cache.

\paragraph{Metrics.}

We evaluate translation quality using COMET~\citep{rei2020comet}, chrF++~\citep{popovic2017chrf}, and SacreBLEU~\citep{post2018sacrebleu}. We use \texttt{Unbabel/wmt22-comet-da} for COMET and report scores on a $0$--$100$ scale; we set the word order to 2 for chrF++; SacreBLEU uses the \texttt{zh} tokenizer for Chinese targets and \texttt{13a} for English targets. Our primary efficiency metric is mean model-only sentence-level generation latency on the GPU at batch size 1. For the LLaDA family, we report the number of function evaluations (NFE). Because tokenizers differ across models, we do not use cross-model token/s as the primary speed metric.

\paragraph{Implementation Details.}

We run all experiments on a single NVIDIA RTX 3090 using BF16 and batch size 1. Autoregressive models stop at their native EOS token, with a safety limit of 256 new tokens. Vanilla LLaDA uses a 128-token canvas, 128 denoising steps, and blocks of length 32. Fast-dLLM and TRAC both use a prefix KV cache, a block length of 32, a maximum canvas length of 128, and a temperature of 0. TRAC uses $r_{\mathrm{en\rightarrow zh}}=0.8$, $r_{\mathrm{zh\rightarrow en}}=1.35$, $\alpha=0.35$, $\tau_{\mathrm{high}}=0.8$, $\tau_{\mathrm{low}}=0.5$, $\epsilon=0.04$, $k=2$, and $K=8$.

All parameters remain fixed across test sets. Latency includes model prefill and generation, but excludes prompt construction, tokenization, CPU-to-GPU transfer, detokenization, file writing, and metric computation.  We reset peak memory statistics before each configuration. For each configuration, the reported latency is the per-sentence mean from one complete benchmark run; model loading time is excluded.

\subsection{Main Results}
Table~\ref{tab:main-results} reports translation quality and inference efficiency across all six dataset--direction settings, together with macro averages.

Compared with Fast-dLLM, TRAC reduces model-only sentence-level generation latency in all six settings. Macro-average latency falls from $1.145$\,s to $0.573$\,s, corresponding to an additional $2.00\times$ speedup, while COMET changes only from $82.740$ to $82.720$. Speedup ranges from $1.91\times$ to $2.19\times$ across settings. TRAC also reduces average NFE from $22.78$ to $16.01$, a $29.7\%$ reduction. Compared with Vanilla LLaDA, TRAC reduces average latency from $9.014$\,s to $0.573$\,s, yielding a $15.72\times$ speedup, while NFE falls from $128$ to $16.01$. These wall-clock gains are therefore accompanied by substantial reductions in denoising computation.

\begin{figure}[t]
    \centering
    \includegraphics[width=\columnwidth,height=0.19\textheight,keepaspectratio]{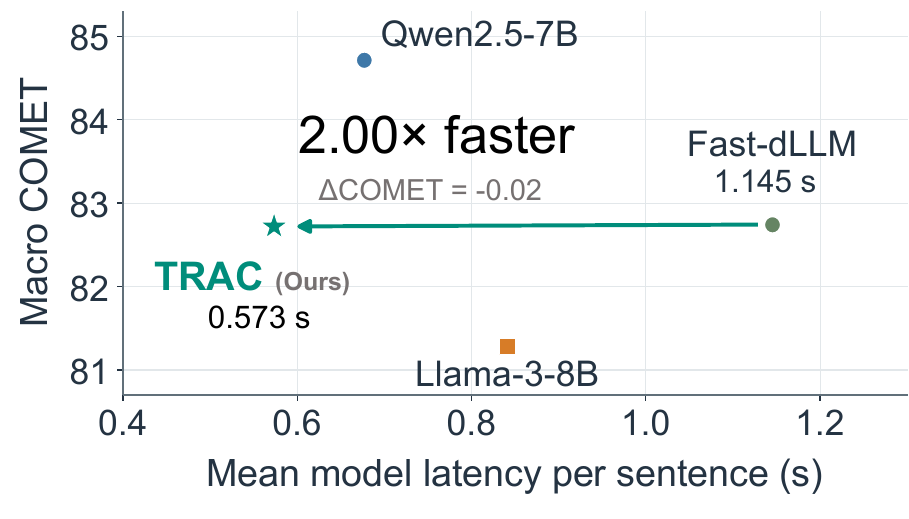}
    \caption{Macro-averaged quality and latency across the six dataset--direction settings.  }
    \label{fig:quality-latency}
\end{figure}

The autoregressive baselines Qwen2.5-7B-Instruct and Llama-3-8B-Instruct have average latencies of $0.677$\,s and $0.841$\,s, respectively, compared with $0.573$\,s for TRAC. TRAC therefore achieves lower average sentence-level latency, although Qwen still has higher macro-average COMET ($84.714$ vs.\ $82.720$). Overall, TRAC substantially improves the quality--latency trade-off of LLaDA-based translation, with a relatively small translation-quality gap to autoregressive models.

\subsection{Ablations and Analysis}

To assess the contributions of the two core modules, we add Translation-Aware Adaptive Canvas and Temporal Commitment to Fast-dLLM individually and in combination. Table~\ref{tab:ablation} reports macro-averaged results across the six test settings.

\begin{table}[!h]
\centering
\small
\setlength{\tabcolsep}{4.4pt}
\begin{tabular}{lrrr}
\toprule
Method & COMET $\uparrow$ & Latency (s) $\downarrow$ & NFE $\downarrow$ \\
\midrule
Fast-dLLM & 82.74 & 1.145 & 22.8 \\
Canvas only & \textbf{83.52} & 0.711 & 20.5 \\
Temporal only & 82.35 & 0.922 & 18.4 \\
\method{} & 82.72 & \textbf{0.573} & \textbf{16.0} \\
\bottomrule
\end{tabular}
\caption{Macro-average results for ablations of the core modules. }
\label{tab:ablation}
\end{table}

\paragraph{Effect of Translation-Aware Adaptive Canvas.}

Adding Adaptive Canvas alone reduces average latency from $1.145$\,s to $0.711$\,s, yielding a $1.61\times$ speedup, while NFE decreases from $22.78$ to $20.51$. This gain comes primarily from reducing the target space while retaining the Fast-dLLM token commitment rule. The mean initial canvas length is $40.56$ tokens, and only about $1.86\%$ of examples trigger fallback, indicating that most translation instances do not require the fixed maximum workspace of 128 tokens. This configuration also improves macro-average COMET from $82.740$ to $83.516$.



\paragraph{Effect of Temporal Commitment.}
With the canvas fixed at 128 tokens, Temporal Commitment reduces average NFE from 22.78 to 18.38 and latency from 1.145 s to 0.922 s, yielding a 1.24$\times$ speedup. This reduction comes from allowing medium-confidence positions that have already become temporally stable to be committed before they reach the original high-confidence threshold, thereby avoiding unnecessary refinement in later denoising steps. Its macro-average COMET is 82.349, reflecting a modest quality--efficiency trade-off introduced by more aggressive early commitment. 

\paragraph{Complementarity.}

The full TRAC framework achieves the lowest average latency ($0.573$\,s) and NFE ($16.01$). After Adaptive Canvas reduces latency to $0.711$\,s, Temporal Commitment provides a further reduction of about $19\%$. The two modules thus address complementary dimensions of computation: the former reduces the target space processed in each forward pass, while the latter reduces the number of refinement steps for the allocated positions.

\FloatBarrier
\section{Conclusion}
\label{sec:conclusion}

We study the underutilized potential of parallel decoding in diffusion language models for low-latency machine translation. Our analysis of Fast-dLLM decoding identifies two complementary forms of computational redundancy in diffusion-based translation: target-space redundancy caused by a fixed target canvas and temporal refinement redundancy caused by commitment rules based on instantaneous confidence.

Motivated by these observations, we propose TRAC, a training-free framework for adaptive diffusion-based translation decoding. Translation-Aware Adaptive Canvas allocates target space dynamically according to source sentence length and translation direction. Dual-Channel Token Commitment preserves the high-confidence fast path while selectively committing medium-confidence tokens early when they exhibit temporal stability. On bidirectional Chinese--English translation using FLORES, WMT23, and NTREX-128, TRAC achieves an additional speedup of approximately $2.00\times$ in average sentence-level latency over Fast-dLLM while maintaining similar overall translation quality and substantially reducing the number of model forward passes required for diffusion denoising. These results identify translation-aware target-space allocation and token-level stopping policies as two effective ways to further exploit parallel inference in dLLMs.

\clearpage
\section*{Limitations}
\label{sec:limitations}

This work has several limitations. First, our experiments are primarily based on LLaDA-8B-Instruct, and TRAC's generalization to other masked diffusion language models requires further validation. Although our method does not modify model parameters and both modules operate at inference time, confidence distributions, decoding trajectories, and block-wise generation behavior may differ across dLLM backbones.

Second, we focus on En$\rightarrow$Zh and Zh$\rightarrow$En translation. Translation-Aware Adaptive Canvas relies on direction-specific source--target length relationships. Its applicability to other language pairs, particularly languages with substantially different morphology and length-compression ratios, therefore requires further evaluation. We select the direction-specific ratio $r_d$ and several temporal commitment thresholds on a development set; adaptive parameter strategies that require no development-set tuning remain unexplored.

Finally, Dual-Channel Token Commitment approximates the cross-step Jensen--Shannon divergence over the full vocabulary using a sparse Top-$K$+OTHER distribution. This approximation substantially reduces the storage cost of historical states but cannot fully preserve changes in the probabilities of individual tokens in the vocabulary tail. Its stability estimates may therefore deviate from those based on full-vocabulary JSD.

\bibliography{custom}

\appendix

\section{Full Results}
\label{sec:full-results}
Table~\ref{tab:full-results} presents results for each of the three benchmarks and two translation directions. 

\begin{table*}[t]
\centering
\scriptsize
\setlength{\tabcolsep}{2.2pt}
\begin{minipage}[t]{0.485\textwidth}
\centering
\resizebox{\linewidth}{!}{\begin{tabular}{llrrrrr}
\toprule
Data & Method & COMET & chrF++ & BLEU & Lat. & NFE \\
\midrule
\multirow{5}{*}{FLORES} & Qwen & 87.62 & 26.76 & 41.18 & 0.646 & -- \\
& Llama & 82.08 & 19.26 & 28.50 & 0.892 & -- \\
& Vanilla & 86.54 & 26.98 & 39.86 & 9.235 & 128.00 \\
& Fast-dLLM & 85.97 & 25.53 & 37.51 & 1.232 & 24.42 \\
& TRAC & 85.21 & 25.05 & 37.68 & \textbf{0.616} & 17.46 \\
\midrule
\multirow{5}{*}{WMT23} & Qwen & 85.00 & 28.93 & 41.64 & \textbf{0.517} & -- \\
& Llama & 79.74 & 23.38 & 32.91 & 0.681 & -- \\
& Vanilla & 84.34 & 31.15 & 44.51 & 8.712 & 128.00 \\
& Fast-dLLM & 83.40 & 28.77 & 40.87 & 1.062 & 21.06 \\
& TRAC & 83.42 & 29.02 & 40.98 & 0.536 & 15.36 \\
\midrule
\multirow{5}{*}{NTREX} & Qwen & 84.93 & 22.38 & 34.03 & \textbf{0.671} & -- \\
& Llama & 78.98 & 16.65 & 23.36 & 0.914 & -- \\
& Vanilla & 83.01 & 21.58 & 31.06 & 9.244 & 128.00 \\
& Fast-dLLM & 82.12 & 20.66 & 29.03 & 1.321 & 26.26 \\
& TRAC & 81.96 & 20.27 & 29.15 & 0.673 & 19.71 \\
\bottomrule
\end{tabular}}

\smallskip
(a) English$\rightarrow$Chinese
\end{minipage}\hfill
\begin{minipage}[t]{0.485\textwidth}
\centering
\resizebox{\linewidth}{!}{\begin{tabular}{llrrrrr}
\toprule
Data & Method & COMET & chrF++ & BLEU & Lat. & NFE \\
\midrule
\multirow{5}{*}{FLORES} & Qwen & 86.80 & 55.19 & 27.59 & 0.706 & -- \\
& Llama & 85.57 & 52.16 & 22.40 & 0.830 & -- \\
& Vanilla & 84.87 & 53.40 & 27.41 & 8.905 & 128.00 \\
& Fast-dLLM & 84.69 & 52.89 & 26.62 & 0.973 & 19.10 \\
& TRAC & 84.98 & 53.56 & 26.23 & \textbf{0.444} & 12.08 \\
\midrule
\multirow{5}{*}{WMT23} & Qwen & 79.74 & 47.05 & 22.01 & 0.805 & -- \\
& Llama & 78.62 & 46.05 & 20.50 & 0.896 & -- \\
& Vanilla & 77.95 & 45.43 & 21.54 & 8.867 & 128.00 \\
& Fast-dLLM & 77.72 & 44.60 & 21.06 & 1.215 & 24.71 \\
& TRAC & 78.57 & 45.24 & 20.75 & \textbf{0.637} & 17.23 \\
\midrule
\multirow{5}{*}{NTREX} & Qwen & 84.19 & 50.37 & 22.03 & 0.716 & -- \\
& Llama & 82.70 & 47.21 & 17.36 & 0.835 & -- \\
& Vanilla & 83.05 & 48.42 & 21.71 & 9.124 & 128.00 \\
& Fast-dLLM & 82.54 & 47.82 & 21.34 & 1.069 & 21.13 \\
& TRAC & 82.19 & 47.90 & 20.08 & \textbf{0.535} & 14.23 \\
\bottomrule
\end{tabular}}

\smallskip
(b) Chinese$\rightarrow$English
\end{minipage}
\caption{Detailed main results. Latency is measured in seconds per sentence.}
\label{tab:full-results}
\label{tab:full-results-en-zh}
\label{tab:full-results-zh-en}
\end{table*}

\section{Reproducibility and Implementation Details}
\label{sec:appendix-protocol}

All speed measurements use the same model inference path. Text is saved and quality metrics are computed outside this path; the timed path includes neither I/O nor evaluation-model calls. We run on a single NVIDIA RTX 3090 with BF16, batch size 1, and random seed 42. For each configuration, we warm up on 20 development examples and perform CUDA synchronization at the timing boundaries. Latency includes model prefill and generation, but excludes prompt construction, tokenization, CPU--GPU transfer, detokenization, file writing, metric computation, and model loading. For FLORES, we combine dev and devtest (2,009 sentences per direction). WMT23 contains 2,074 En$\rightarrow$Zh and 1,976 Zh$\rightarrow$En examples, and NTREX-128 contains 1,997 sentences per direction.




\section{Per-Setting Component Ablations}
\label{sec:appendix-ablation}

Table~\ref{tab:ablation} in the main text reports macro-averaged ablation results. Table~\ref{tab:appendix-ablation-en-zh}(a) and Table~\ref{tab:appendix-ablation-zh-en}(b) show the full results for the two translation directions, respectively.

Apart from the module being evaluated, all ablations use the same LLaDA-8B-Instruct weights, translation prompt, prefix KV cache, block size, and Fast-dLLM high-confidence threshold. \textit{Canvas} enables only Translation-Aware Adaptive Canvas with the same conservative fallback. \textit{Temporal} retains a fixed 128-token canvas and enables only dual-channel commitment. The two single-module configurations therefore isolate the effects of target-space allocation and commitment based on temporal stability, respectively; full TRAC combines both. Latency and NFE include the cost of fallback retries in every setting.

\begin{table*}[t]
\centering
\scriptsize
\setlength{\tabcolsep}{2.2pt}
\begin{minipage}[t]{0.485\textwidth}
\centering
\resizebox{\linewidth}{!}{\begin{tabular}{llrrrrr}
\toprule
Data & Method & COMET & Lat. & NFE & $C_0$ & FB (\%) \\
\midrule
\multirow{4}{*}{FLORES} & Fast & 85.97 & 1.232 & 24.42 & 128.00 & -- \\
& Canvas & 86.44 & 0.740 & 22.03 & 34.36 & 0.25 \\
& Temporal & 85.10 & 0.992 & 19.58 & 128.00 & -- \\
& TRAC & 85.21 & \textbf{0.616} & \textbf{17.46} & 34.36 & 0.25 \\
\midrule
\multirow{4}{*}{WMT23} & Fast & 83.40 & 1.062 & 21.06 & 128.00 & -- \\
& Canvas & 84.20 & 0.652 & 18.82 & 34.85 & 1.45 \\
& Temporal & 83.07 & 0.874 & 17.60 & 128.00 & -- \\
& TRAC & 83.42 & \textbf{0.536} & \textbf{15.36} & 34.85 & 0.63 \\
\midrule
\multirow{4}{*}{NTREX} & Fast & 82.12 & 1.321 & 26.26 & 128.00 & -- \\
& Canvas & 82.73 & 0.783 & 23.53 & 36.60 & 1.00 \\
& Temporal & 81.53 & 1.102 & 22.18 & 128.00 & -- \\
& TRAC & 81.96 & \textbf{0.673} & \textbf{19.71} & 36.60 & 0.90 \\
\bottomrule
\end{tabular}}

\smallskip
(a) English$\rightarrow$Chinese
\end{minipage}\hfill
\begin{minipage}[t]{0.485\textwidth}
\centering
\resizebox{\linewidth}{!}{\begin{tabular}{llrrrrr}
\toprule
Data & Method & COMET & Lat. & NFE & $C_0$ & FB (\%) \\
\midrule
\multirow{4}{*}{FLORES} & Fast & 84.69 & 0.973 & 19.10 & 128.00 & -- \\
& Canvas & 85.63 & 0.584 & 16.83 & 44.55 & 2.09 \\
& Temporal & 84.59 & 0.749 & 14.67 & 128.00 & -- \\
& TRAC & 84.98 & \textbf{0.444} & \textbf{12.08} & 44.55 & 0.75 \\
\midrule
\multirow{4}{*}{WMT23} & Fast & 77.72 & 1.215 & 24.71 & 128.00 & -- \\
& Canvas & 78.77 & 0.827 & 22.76 & 46.32 & 3.85 \\
& Temporal & 78.03 & 0.970 & 19.77 & 128.00 & -- \\
& TRAC & 78.57 & \textbf{0.637} & \textbf{17.23} & 46.32 & 1.92 \\
\midrule
\multirow{4}{*}{NTREX} & Fast & 82.54 & 1.069 & 21.13 & 128.00 & -- \\
& Canvas & 83.32 & 0.679 & 19.08 & 46.69 & 2.50 \\
& Temporal & 81.77 & 0.844 & 16.47 & 128.00 & -- \\
& TRAC & 82.19 & \textbf{0.535} & \textbf{14.23} & 46.69 & 1.35 \\
\bottomrule
\end{tabular}}

\smallskip
(b) Chinese$\rightarrow$English
\end{minipage}
\caption{Per-setting component ablations. \textit{Canvas} denotes translation-aware canvas allocation only; \textit{Temporal} denotes dual-channel commitment only; FB is the fallback rate.}
\label{tab:appendix-ablation}
\label{tab:appendix-ablation-en-zh}
\label{tab:appendix-ablation-zh-en}
\end{table*}

\section{Detailed Decoding Diagnostics}
\label{sec:appendix-diagnostics}

\subsection{Canvas Utilization and Fallback}
\label{sec:canvas-diagnostics}

\begin{figure*}[t]
    \centering
    \includegraphics[width=0.78\textwidth]{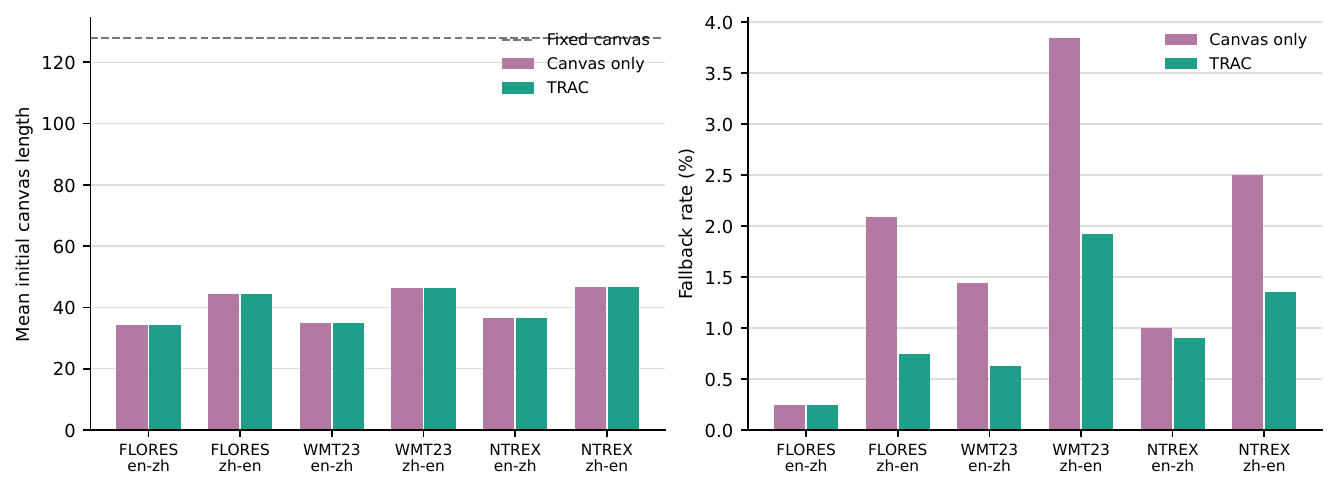}
\caption{Per-setting adaptive canvas statistics. Left: mean initial canvas length for Canvas only and TRAC; the dashed line indicates a fixed 128-token canvas. Right: fallback rates for the two configurations.}
\label{fig:canvas-diagnostics}
\end{figure*}

\begin{table*}[!h]
\centering
\small
\setlength{\tabcolsep}{10pt}
\resizebox{0.82\textwidth}{!}{%
\begin{tabular}{lrrrr}
\toprule
Acceptance rule & Stable precision (95\% CI) & Stable recall & Coverage & Earlier NFE \\
\midrule
Medium confidence only & 76.79 [75.36, 78.22] & 100.00 & 100.00 & 1.94 \\
+ top-1 persistence & 79.60 [77.70, 81.57] & 51.90 & 50.07 & 1.91 \\
+ persistence + sparse JSD & 79.82 [77.78, 81.93] & 46.59 & 44.82 & 1.96 \\
\bottomrule
\end{tabular}}
\caption{Temporal commitment audit of frozen Fast-dLLM trajectories (400 FLORES sentences, 200 per direction). A positive label indicates that the current top-1 prediction remains unchanged in subsequent masked observations and matches the token ultimately committed by Fast-dLLM. Confidence intervals use 2,000 sentence-level bootstrap resamples stratified by translation direction.}
\label{tab:appendix-temporal-audit}
\end{table*}

Figure~\ref{fig:canvas-diagnostics} reports per-setting statistics for dynamic canvas allocation and conservative fallback. Both Canvas only and full TRAC restrict the initial canvas to a compact set of block-aligned candidates. Fallback is triggered only for a small number of empty or severely short outputs.

We define canvas utilization as $L_{\mathrm{out}}/128$, where $L_{\mathrm{out}}$ counts the valid LLaDA output tokens before the first EOS token, excluding EOS itself and all post-EOS positions. We compute this statistic from the complete outputs for all six test settings (12,062 sentences in total), without reference translations. It represents the fraction of positions in a fixed canvas that ultimately contain the valid translation and should not be interpreted as an exact FLOPs ratio.

The initial canvas length is computed from the direction-specific ratio $r_d$ and the source sentence's token length. The length estimate is first rounded to an integer and clipped to $[32,128]$, then aligned upward to a multiple of 32 tokens. This discrete alignment keeps the dynamic canvas compatible with Fast-dLLM's block-wise prefix KV cache. The fallback criterion examines only the valid output from the first decoding attempt: if the output is empty or its length is less than $\alpha=0.35$ times the predicted length $\hat L_d$, TRAC retries from the initial state with $C_{\max}=128$. This mechanism uses neither reference lengths nor an additional quality model.

\subsection{Temporal Commitment Audit}

We conduct a non-interventional audit of frozen Fast-dLLM trajectories on a fixed set of 400 FLORES examples stratified by reference length, with 200 sentences per direction. The analysis excludes EOS and post-EOS positions, and candidate observations satisfy $0.5\leq c_i^{(t)}<0.8$. Model outputs, NFE, and commitment order remain identical to those of the original Fast-dLLM. 

Table~\ref{tab:appendix-temporal-audit} shows that adding persistence and sparse JSD to acceptance based solely on medium confidence increases trajectory-stable precision from $76.79\%$ to $79.82\%$ while retaining $44.82\%$ of medium-confidence observations. On average, positions first satisfy the rule $1.96$ NFE before their eventual commitment by Fast-dLLM, indicating an opportunity for earlier commitment.

\NewDocumentEnvironment{appendixtabular}{m m +b}{%
  \begingroup
  \setbox0=\hbox{\begin{tabular}{#1}#3\end{tabular}}%
  \setlength{\tabcolsep}{\dimexpr\tabcolsep+(\textwidth-\wd0)/#2\relax}%
  \begin{tabular}{#1}#3\end{tabular}%
  \endgroup
}{}

\begin{table*}[t]
\centering
\captionsetup{singlelinecheck=false}
\footnotesize
\setlength{\tabcolsep}{3pt}
\renewcommand{\arraystretch}{1.25}

\begin{appendixtabular}{lrrrrrr}{14}
\toprule
& \multicolumn{3}{c}{En$\rightarrow$Zh}
& \multicolumn{3}{c}{Zh$\rightarrow$En} \\

\cmidrule(lr){2-4}
\cmidrule(lr){5-7}

Allocation policy
& COMET $\uparrow$
& Lat. $\downarrow$
& $\Delta$COMET
& COMET $\uparrow$
& Lat. $\downarrow$
& $\Delta$COMET \\

\midrule

Fixed-32
& 82.69 & \textbf{0.809} & -1.52
& 75.74 & \textbf{0.673} & -4.38 \\

Fixed-64
& 83.99 & 1.058 & -0.22
& 79.31 & 0.887 & -0.81 \\

Fixed-96
& 84.56 & 1.424 & +0.35
& 79.79 & 1.248 & -0.33 \\

Fixed-128
& 84.21 & 1.683 & +0.00
& 80.12 & 1.481 & +0.00 \\

\rowcolor[gray]{0.94}
Adaptive canvas (ours)
& \textbf{85.08} & 1.048 & +0.87
& \textbf{80.96} & 1.026 & +0.84 \\

\bottomrule
\end{appendixtabular}

\caption{
Canvas policies at confidence 0.8.
$\Delta$COMET is relative to fixed-128.
}

\label{tab:dev-fixed-canvas}
\end{table*}

\begin{table*}[!t]
\centering
\captionsetup{singlelinecheck=false}
\footnotesize
\setlength{\tabcolsep}{3pt}
\renewcommand{\arraystretch}{1.25}

\begin{appendixtabular}{lrrrrrr}{14}
\toprule

& \multicolumn{3}{c}{En$\rightarrow$Zh}
& \multicolumn{3}{c}{Zh$\rightarrow$En} \\

\cmidrule(lr){2-4}
\cmidrule(lr){5-7}

Commitment policy
& COMET $\uparrow$
& Lat. $\downarrow$
& NFE $\downarrow$
& COMET $\uparrow$
& Lat. $\downarrow$
& NFE $\downarrow$ \\

\midrule

Confidence-only, $\tau=0.5$
& 81.90 & \textbf{0.656} & \textbf{18.65}
& 78.05 & \textbf{0.523} & \textbf{13.97} \\

Confidence-only, $\tau=0.6$
& 83.57 & 0.800 & 23.00
& 79.89 & 0.706 & 18.93 \\

Confidence-only, $\tau=0.7$
& 84.50 & 0.961 & 27.05
& 80.96 & 0.859 & 23.66 \\

Confidence-only, $\tau=0.8$
& 85.08 & 1.048 & 30.50
& 80.96 & 1.026 & 28.51 \\

Confidence-only, $\tau=0.9$
& \textbf{85.15} & 1.152 & 33.76
& 80.89 & 1.243 & 34.15 \\

\midrule

\rowcolor[gray]{0.94}
TRAC, $\tau_{\rm low}=0.5$ (original)
& 84.35 & 0.901 & 25.27
& 80.23 & 0.817 & 21.93 \\

\rowcolor[gray]{0.94}
TRAC, $\tau_{\rm low}=0.6$
& 85.03 & 0.996 & 28.03
& 80.89 & 0.939 & 25.48 \\

\rowcolor[gray]{0.94}
TRAC, $\tau_{\rm low}=0.7$
& 85.01 & 1.056 & 29.82
& \textbf{80.99} & 1.031 & 27.49 \\

\bottomrule
\end{appendixtabular}

\caption{
Commitment policies under the same adaptive canvas.
TRAC fixes high confidence at 0.8;
0.5 is the original lower threshold.
}

\label{tab:dev-threshold}
\end{table*}

In this audit, trajectory-stable is a decoding-trajectory label rather than a measure of correctness against the reference translation. It requires a position's current top-1 prediction to remain unchanged in subsequent observations while the position is masked and to match the token ultimately committed by frozen Fast-dLLM. To avoid treating multiple positions in the same long sentence as independent samples, we compute the 95\% confidence intervals in the table using 2,000 sentence-level bootstrap resamples stratified by translation direction. This diagnostic evaluates only the evidence for acceptance through the second channel; TRAC's actual quality and latency are measured by running its full counterfactual decoding procedure in the main experiments.

\balance
\section{Supplementary Experiments}

\label{sec:appendix-dev}

\subsection{Evaluation Protocol}

For the supplementary experiments in this section, all configurations use 200 sentences per direction from source-length-stratified subsets, LLaDA-8B-Instruct, common prompts, BF16, batch size one and block size 32. Latency includes retries/probes and is descriptive; NFE counts actual backbone calls.

 \begin{table*}[!t]
\centering
\captionsetup{singlelinecheck=false}
\footnotesize
\setlength{\tabcolsep}{3pt}
\renewcommand{\arraystretch}{1.25}

\begin{appendixtabular}{clrrrrrr}{16}
\toprule

& &
\multicolumn{3}{c}{En$\rightarrow$Zh}
& \multicolumn{3}{c}{Zh$\rightarrow$En} \\

\cmidrule(lr){3-5}
\cmidrule(lr){6-8}

$\tau_{\rm low}$
& Rule
& COMET
& Lat.
& $\Delta$COMET [95\% CI]
& COMET
& Lat.
& $\Delta$COMET [95\% CI] \\

\midrule

0.5 & Persistence
& 84.27 & \textbf{0.863} & --
& 80.05 & \textbf{0.767} & -- \\

\rowcolor[gray]{0.94}
0.5 & + JSD (TRAC)
& \textbf{84.35} & 0.901
& $+0.084\ [-0.126,+0.319]$
& \textbf{80.23} & 0.817
& $+0.174\ [-0.086,+0.416]$ \\

\midrule

0.6 & Persistence
& 84.90 & \textbf{0.963} & --
& 80.79 & \textbf{0.905} & -- \\

\rowcolor[gray]{0.94}
0.6 & + JSD (TRAC)
& \textbf{85.03} & 0.996
& $+0.125\ [+0.007,+0.294]$
& \textbf{80.89} & 0.939
& $+0.096\ [-0.040,+0.329]$ \\

\midrule

0.7 & Persistence
& \textbf{85.01} & \textbf{1.028} & --
& \textbf{80.99} & \textbf{0.985} & -- \\

\rowcolor[gray]{0.94}
0.7 & + JSD (TRAC)
& \textbf{85.01} & 1.056
& $+0.000$ [--]
& \textbf{80.99} & 1.031
& $+0.000$ [--] \\

\bottomrule
\end{appendixtabular}

\caption{
Matched JSD ablation.
Differences and 2,000-resample paired intervals compare
JSD with persistence;
-- indicates an uncomputed interval.
}

\label{tab:dev-temporal}
\end{table*}

\subsection{Adaptive Allocation versus a Fixed Short Canvas}
\label{sec:dev-fixed}

Table~\ref{tab:dev-fixed-canvas} compares adaptive allocation with fixed canvases. Relative to fixed-128, adaptive allocation gains 0.87/0.84 observed COMET in En$\rightarrow$Zh/Zh$\rightarrow$En with 37.7\%/30.7\% lower recorded latency. Fixed-32 is fastest but loses 1.52/4.38 COMET. Compared with fixed-64, adaptive allocation gains 1.09 COMET at similar En$\rightarrow$Zh latency and 1.64 COMET at higher Zh$\rightarrow$En latency. Its advantage is balancing quality and workspace cost.

\begin{table*}[!t]
\centering
\captionsetup{singlelinecheck=false}
\footnotesize
\setlength{\tabcolsep}{3pt}
\renewcommand{\arraystretch}{1.25}

\begin{appendixtabular}{lrrrrrr}{14}
\toprule

& \multicolumn{3}{c}{En$\rightarrow$Zh}
& \multicolumn{3}{c}{Zh$\rightarrow$En} \\

\cmidrule(lr){2-4}
\cmidrule(lr){5-7}

Adaptive decoding policy
& COMET $\uparrow$
& Lat. $\downarrow$
& NFE $\downarrow$
& COMET $\uparrow$
& Lat. $\downarrow$
& NFE $\downarrow$ \\

\midrule

LESS + cache, $\tau=0.5$
& 83.47 & 0.989 & 27.98
& 79.67 & 0.935 & 25.28 \\

LESS + cache, $\tau=0.6$
& 84.76 & 1.109 & 31.04
& 80.47 & 1.114 & 30.22 \\

LESS + cache, $\tau=0.75$
& \textbf{85.15} & 1.208 & 34.56
& 80.87 & 1.292 & 35.17 \\

LESS + cache, $\tau=0.8$
& 85.13 & 1.273 & 35.53
& 80.95 & 1.347 & 36.91 \\

\midrule

EV, aligned candidates
& 84.82 & 1.130 & 31.50
& \textbf{80.99} & 1.114 & 30.00 \\

EV, raw then align
& 85.08 & 1.271 & 35.38
& 80.74 & 1.213 & 32.73 \\

\midrule

Confidence-only, $\tau=0.8$
& 85.08 & 1.048 & 30.50
& 80.96 & 1.026 & 28.51 \\

\midrule

\rowcolor[gray]{0.94}
TRAC, $\tau_{\rm low}=0.5$ (original)
& 84.35 & \textbf{0.901} & \textbf{25.27}
& 80.23 & \textbf{0.817} & \textbf{21.93} \\

\rowcolor[gray]{0.94}
TRAC, $\tau_{\rm low}=0.6$
& 85.03 & 0.996 & 28.03
& 80.89 & 0.939 & 25.48 \\

\rowcolor[gray]{0.94}
TRAC, $\tau_{\rm low}=0.7$
& 85.01 & 1.056 & 29.82
& \textbf{80.99} & 1.031 & 27.49 \\

\bottomrule
\end{appendixtabular}

\caption{
All adaptive-cache LESS and EV selector settings,
with common reference policies.
Gray: TRAC.
Bold: column best at displayed precision, including ties.
Latency includes probes and retries.
}

\label{tab:dev-related-adaptive}
\end{table*}

\subsection{Temporal Commitment versus Confidence Tuning}
\label{sec:dev-threshold}

Table~\ref{tab:dev-threshold} compares temporal commitment with confidence tuning. Original TRAC uses 17.1\%/23.1\% fewer calls than confidence-only 0.8, with a 0.73-point COMET decrease. Raising its lower threshold to 0.6 brings observed COMET within 0.07 points of that baseline, with 8.1\%/10.6\% fewer calls and 5.0\%/8.5\% lower recorded latency. Temporal commitment thus offers tunable quality--cost settings. 

\subsection{Incremental Role of Distributional Stability}
\label{sec:dev-temporal}

Table~\ref{tab:dev-temporal} removes JSD at matched canvas, fallback and persistence settings. At lower thresholds 0.5/0.6, JSD yields small observed COMET increases in both directions with 3.5\%--6.5\% higher recorded latency; at 0.7, COMET is unchanged. Its observed contribution is therefore concentrated at the more permissive thresholds. The distributional constraint provides a more conservative acceptance rule and adjusts the quality--efficiency trade-off.

\subsection{Comparison with Related Adaptive Policies}
\label{sec:dev-related}

Table~\ref{tab:dev-related-adaptive} compares cache/selector adaptations on common Instruct weights. Original TRAC has the lowest NFE and recorded latency in this table, while some alternatives achieve higher COMET. At 0.6, TRAC uses 18.9\%/27.5\% fewer calls than adaptive-cache LESS at 0.75, with COMET differences of $-0.13$/$+0.02$. At 0.7 it reaches 85.01/80.99 COMET versus EV's 84.82/80.99, with 5.3\%/8.4\% fewer calls. TRAC thus offers reduced computation at competitive observed quality without candidate-canvas probes. 

\end{document}